\documentclass[letterpaper, 10pt, journal, twoside]{ieeetran}
\usepackage[utf8]{inputenc}

\IEEEoverridecommandlockouts

\usepackage{cite}
\usepackage{ulem}
\usepackage{xcolor}
\usepackage{acronym}
\usepackage{caption}
\usepackage{leftidx}
\usepackage{listings}
\usepackage{graphicx}
\usepackage{textcomp}
\usepackage{multirow}
\usepackage{booktabs}
\usepackage{colortbl}
\usepackage{subcaption}
\usepackage{amsmath,amssymb,amsfonts}

\usepackage{booktabs}
\usepackage{array}
\usepackage{savesym}
\savesymbol{checkmark}
\usepackage{dingbat}
\usepackage{bbding}
\usepackage{colortbl}
\usepackage[table]{xcolor}
\usepackage{csquotes}
\usepackage{soul}
\usepackage{svg}

\usepackage{algorithm}      
\usepackage{algpseudocode}  
\newcommand{\change}[2]{\textcolor{black}{#2}}
\newcommand{\add}[1]{\textcolor{black}{#1}}

\newcommand{\changetwo}[2]{\textcolor{black}{#2}}
\newcommand{\addtwo}[1]{\textcolor{black}{#1}}

\usepackage{hyperref}
\usepackage{cleveref}

\acrodef{TP}{True Positive}
\acrodef{FP}{False Positive}
\acrodef{FN}{False Negative}
\acrodef{VSLAM}{Visual SLAM}
\acrodef{STD}{Standard Deviation}
\acrodef{ROS}{Robot Operating System}
\acrodef{RMSE}{Root Mean Square Error}
\acrodef{ATE}{Absolute Trajectory Error}
\acrodef{RANSAC}{RANdom SAmple Consensus}
\acrodef{CNN}{Convolutional Neural Network}
\acrodef{LiDAR}{Light Detection And Ranging}
\acrodef{SLAM}{Simultaneous Localization and Mapping}

\newcommand{\vgraphs}{\textit{vS-Graphs}}
\newcommand{\ivgraphs}{\textit{ivS-Graphs}}

\definecolor{red}{HTML}{fd8f8f}
\definecolor{greend}{HTML}{57e377}
\definecolor{greenl}{HTML}{b8fb8a}
\definecolor{lyellow}{HTML}{fefdb4}
\definecolor{orange}{HTML}{ffd5ab}

\colorlet{red}{red!50}
\colorlet{yellow}{yellow!50}
\colorlet{greenl}{greenl!50}
\colorlet{greend}{greend!50}

\title{\LARGE \bf BIM-Informed Visual SLAM for Construction Environments}

\author{
    Asier Bikandi-Noya$^{1}$, Miguel Fernandez-Cortizas$^{1}$, Muhammad Shaheer$^{1}$,\\ Ali Tourani$^{1}$, Holger Voos$^{1}$, and Jose Luis Sanchez-Lopez$^{1}$  

    \thanks{Manuscript received: May 31, 2026; Revised: July 29, 2026; Accepted: September 7, 2026.}
    \thanks{This paper was recommended for publication by Ayoung Kim upon evaluation of the Associate Editor and Reviewers’ comments.}
    \thanks{$^{1}$Authors are with the Automation and Robotics Research Group, Interdisciplinary Centre for Security, Reliability and Trust (SnT), University of Luxembourg. Holger Voos is also associated with the Faculty of Science, Technology and Medicine, University of Luxembourg, Luxembourg.
    \tt{\small{\{asier.bikandi, muhammad.shaheer, miguel.fernandez, ali.tourani, holger.voos, joseluis.sanchezlopez\}}@uni.lu}}
    \thanks{*
    This work was partially funded by the Fonds National de la Recherche of Luxembourg (FNR) under the project C22/IS/17387634/DEUS and BRIDGES/2025-1/IS/19685965/BARCODE}
    \thanks{*
    We would like to thank CLK S.A.R.L for granting us access to test this work in their construction sites.
    }
    \thanks{*
    For the purpose of Open Access, and in fulfillment of the obligations arising from the grant agreement, the authors have applied a Creative Commons Attribution 4.0 International (CC BY 4.0) license to any Author Accepted Manuscript version arising from this submission.}
} 

\begin{document}

\maketitle

\begin{abstract}

Monitoring building construction sites requires comparing the \textit{as-planned} design with the \textit{as-built} state, which can be estimated in real time using Simultaneous Localization and Mapping (SLAM) \change{to reconstruct the scene and localize the sensor}{techniques}. 
\change{however, visual SLAM, despite its lightweight and cost-effective hardware, is prone to trajectory drift in construction environments due to repetitive layouts, occlusions, and low texture structures.}
{However, visual SLAM is prone to trajectory drift in construction environments, producing maps that are geometrically inaccurate with the actual environment.}
To address this limitation, we augment an existing RGB-D SLAM system with structural priors derived from the Building Information Model (BIM). The system associates detected walls with their BIM counterparts and includes these correspondences as geometric constraints in the back-end optimization, reducing drift and enhancing global consistency.
The proposed method operates in real time and is validated on multiple real construction sites, achieving an average trajectory error reduction of 25.23\% and a 7.14\% improvement in map accuracy over state-of-the-art baselines. Robustness analyses further demonstrate resilience to incomplete BIM data and geometric discrepancies between \textit{as-planned} models and the \textit{as-built} environment.

\noindent \textbf{Video:} \href{https://www.youtube.com/watch?v=jO3rq93ZRtM}{https://www.youtube.com/watch?v=jO3rq93ZRtM}

\noindent Code: \url{https://github.com/snt-arg/BIM_informed_visual_sgraphs}.

\end{abstract}

\vspace{-0.1cm}
\section{Introduction}
\label{sec_intro}
\add{Monitoring the construction of buildings requires estimating the \textit{as-built} state and comparing it with} the \textit{as-planned} design, typically represented as a Building Information Model (BIM)~\cite{bim}, which encodes the structural layout, including walls, floors, and other architectural elements.
Enabling applications such as on-site progress tracking and Augmented Reality (AR) visualization on portable devices~\cite{yigitbas2023supportingAR} requires real-time estimation and continuous updating of the \textit{as-built} state.
\add{Visual Simultaneous Localization and Mapping (Visual SLAM) provides a framework for jointly reconstructing the environment and localizing the sensor~\cite{cadena2016past}.}

\change{However, visual SLAM remains vulnerable to the characteristics of construction environments. Repetitive layouts, large textureless surfaces, and occlusions introduce ambiguous observations that accumulate as trajectory drift, producing maps that are geometrically inaccurate with the actual environment and limit their reliability for monitoring~\cite{sun2025nothingHilti}.}
{However, visual SLAM remains prone to trajectory drift in construction environments, where ambiguous observations accumulate as localization error throughout the trajectory~\cite{sun2025nothingHilti}, producing maps that are geometrically inaccurate with the actual environment, directly compromising the reliability of \textit{as-built} vs.\ \textit{as-planned} comparisons essential for progress tracking.}
\add{Incorporating architectural priors from the BIM into the SLAM pipeline offers the possibility to enforce structural consistency and mitigate drift. Such integration has been investigated in LiDAR-based systems~\cite{isgraphs} that operate in indoor environments where structural elements are already in place and can be matched against the prior. However, to the best of our knowledge, architectural BIM priors have not yet been explicitly embedded within a visual SLAM framework. This integration is non-trivial in the visual domain. Sparser, range-limited depth and higher per-frame noise lead to greater drift accumulation, which in turn complicates the association between detected walls and their BIM counterparts.}

\begin{figure}[t]
    \centering
    \includegraphics[width=0.9\columnwidth]{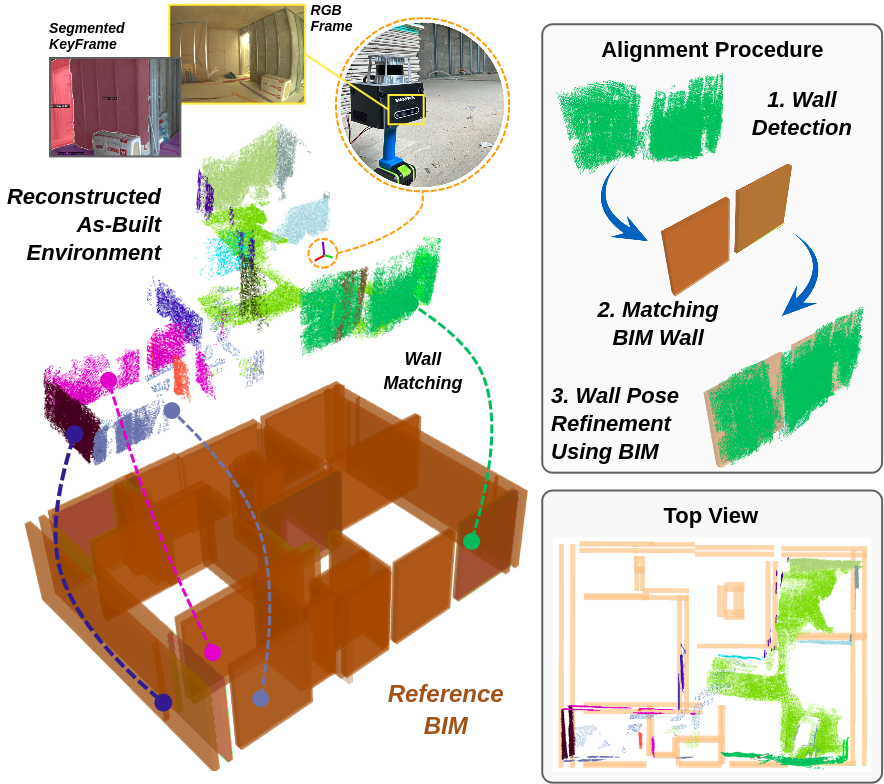}
    \caption{Proposed BIM-informed RGB-D SLAM system. Detected \textit{as-built} walls are matched to their \textit{as-planned} BIM counterparts, with a 2D projection of the alignment.}
    \label{fig:front_image}
\end{figure}

To fill this gap, this work proposes \textit{ivS-Graphs} (illustrated in Fig.~\ref{fig:front_image}), building upon an existing visual SLAM backbone~\cite{tourani2025vsgraphs} that jointly estimates structural planes with keyframe poses. 
\add{ivS-Graphs extends this backbone by introducing BIM walls as prior nodes and matching them against locally detected walls, aligning the evolving map with the \textit{as-planned} design and reducing drift throughout the trajectory.}
\change{Walls are chosen as the key structural element because they are among the earliest and most persistent features in construction environments, providing reliable geometric anchors even under partial observability.}
{\addtwo{The system targets indoor, wall-dominant, building construction environments, such as residential buildings, where walls are among the earliest and most persistent planar features, providing reliable geometric anchors even under partial observability. They are therefore chosen as the key structural element for correspondence with the BIM.}}
The main contributions of this work are:

\begin{itemize}
    \item A novel integration of architectural BIM priors into a visual SLAM framework, reducing trajectory drift by enforcing structural consistency between the \textit{as-built} map and the \textit{as-planned} BIM.
    \item A wall-based initialization and association strategy that establishes correspondences between \textit{as-planned} and \textit{as-built} walls using only two walls as prior information, enabling deployment from the earliest stages of operation.
    \item A system that maintains mapping accuracy under partially built conditions and geometric discrepancies between the \textit{as-planned} and \textit{as-built} models.
\end{itemize}

\begin{figure*}[t]
    \centering
    \includegraphics[width=1.0\textwidth]{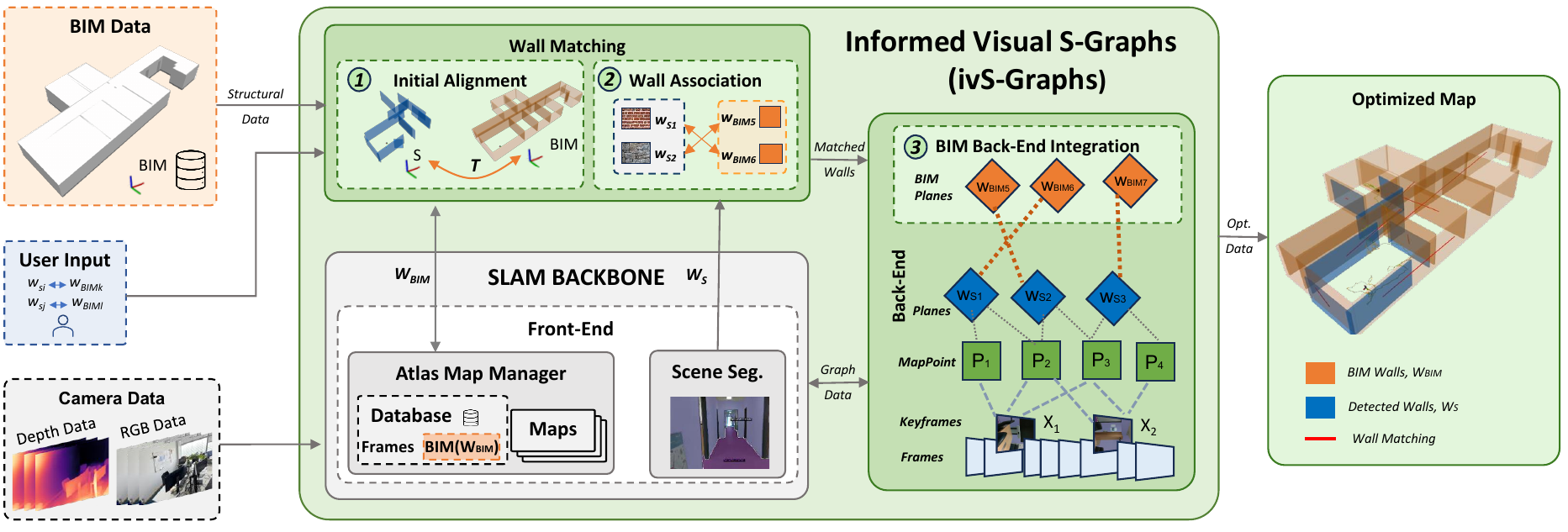} 
\caption{System architecture of ivS-Graphs. The pipeline takes BIM and RGB-D camera data as inputs and BIM walls are extracted offline from the BIM model. The SLAM backbone front-end processes visual data into keyframes, map points, and wall segments. Our contributions are highlighted in green: (\textbf{1}) initial alignment followed by a (\textbf{2}) continuous wall association and (\textbf{3}) back-end graph optimization incorporating wall-to-wall constraints ($\mathbf{W_{\text{BIM}}} \leftrightarrow \mathbf{W_S}$). This constrains camera drift to produce an optimized as-built map (right) aligned with the design.}
 \label{fig:system_architecture}
\end{figure*}

\section{Related Works}
\label{sec_related}

\subsubsection*{\textbf{Visual SLAM in structured environments}}
\textit{ORB-SLAM3}\cite{orb3} provides a widely used feature-based SLAM baseline supporting multiple sensor configurations, and \textit{vS-Graphs}\cite{tourani2025vsgraphs} extends it by integrating structural planes into a factor-graph back-end.
Dense approaches like \textit{BAD-SLAM}~\cite{badslam} offer great geometric detail but often struggle with real-time performance on long construction trajectories. 
More recently, learning-based systems such as DROID-SLAM~\cite{droidslam} and SSF-SLAM~\cite{ssf_slam} have improved robustness through dense bundle adjustment and semantic cues.
However, in construction environments these methods produce drift-affected maps unreliable as an \textit{as-built reference}~\cite{sun2025nothingHilti}.
Since the BIM~\cite{bim} already encodes the expected structural 
layout, it provides a suitable source of architectural priors to 
constrain and correct the SLAM. 

\subsubsection*{\textbf{Global localization with architectural priors}}
A common strategy to integrate prior knowledge, such as BIM or floorplans, in SLAM is to localize the robot within a pre-existing map using particle filters such as Monte Carlo Localization~\cite{dellaert1999monte}. With LiDAR input, this strategy has been extended to use architectural priors for building-scale localization~\cite{blum_precise_2021}.
\add{Further works demonstrate robust pose tracking in CAD drawings~\cite{Boniardi} and semantic BIM-based localization using 2D LiDAR~\cite{Hendrikx}}.
Vision-based methods align monocular RGB detections to 2D floor plans~\cite{zimmerman_constructing_2023} or construct semantic scene graphs conditioned on prebuilt 3D LiDAR maps~\cite{longo2025pixels}. 
While effective for pose estimation, these pipelines localize within a fixed map and do not refine an evolving \textit{as-built} map, often requiring substantial operation for reliable matching.

\subsubsection*{\textbf{Architectural priors in SLAM systems}}
To address these limitations, recent works examine the possibility of coupling BIM with SLAM so that architectural priors directly regularize the trajectory and keep the evolving map consistent with the design. \textit{BIM-SLAM} ~\cite{torres2024bim} integrates BIM into multi-session 3D LiDAR SLAM by seeding a pose-graph from the model, anchoring subsequent sessions to the BIM, and reconstructing new elements not present in the plans. 
Other works combine the \textit{as-built} map and \textit{as-planned} BIM into a joint optimization graph: \textit{iS-Graphs} merges a BIM-derived graph with the online \textit{S-Graph}~\cite{bavle2025s} to enable global localization with room semantics~\cite{isgraphs}, with follow-up work measuring discrepancies between plan and observations~\cite{disgraphs}.
These methods, while benefiting from LiDAR-based high accuracy and $360^\circ$ coverage for room detection, rely on room-level semantics for global localization. This means that in symmetric environments, or when room detection fails or an insufficient number of rooms are observed, these methods cannot establish reliable localization, preventing operation in smaller areas such as a single room.

A few works have extended these ideas to vision-based SLAM. 
For example,~\cite{driftFree} integrates visual SLAM with detailed digital twins to suppress drift; however, \add{this system targets outdoor urban environments and aligns sparse point clouds against a dense photorealistic mesh representing the \textit{as-built} state via ICP. 
This registration is purely geometric, matching points by proximity alone.
Such a dense \textit{as-built} reconstruction must be captured from the environment beforehand, and reflects the clutter and materials that change throughout active construction. Keeping such a prior valid as the site evolves would require its continuous re-capture.}
\add{The \textit{as-planned} BIM available from the design phase offers an alternative: it encodes semantic elements such as walls that do not need to be re-captured as the environment evolves. However, its use in visual SLAM to maintain a drift-bounded \textit{as-built} map remains unexplored, motivating the approach proposed in this work.}

\section{Methodology}
\label{sec_proposed}

\subsection{System Overview}
\label{sec:system_architecture}

\change{Our system builds upon an existing visual SLAM backbone that integrates structural walls into its map representation. Accurate wall estimation is essential for our approach, and for this we leverage RGB-D sensors, which provide immediate metric scale and geometric reliability on the textureless surfaces common in construction sites, where stereo matching often fails~\cite{rgbd_survey}.}
{Our system builds upon an existing visual SLAM backbone that integrates structural walls into its map representation. }
The overall architecture is illustrated in Fig.~\ref{fig:system_architecture}.

\textbf{Inputs.} The system takes three inputs: (i) RGB-D data from the camera, (ii) architectural walls extracted from the BIM, and 
(iii) the correspondences between two BIM walls and their detected counterparts, established through a human interface (Section~III-C).
Both detected and BIM-extracted walls are represented as:
\begin{equation}
w_i = 
\begin{cases}
\mathrm{id}_i, & \text{unique wall identifier} \\
\pi_i = (\mathbf{n}_i, d_i), & \text{supporting plane: normal and distance} \\
\mathbf{c}_i, & \text{centroid position} \\
l_i, t_i, & \text{length and thickness of the wall}
\end{cases}
\end{equation}
\change{Here, $\mathbf{n}_i \in \mathbb{R}^3$ and $d_i \in \mathbb{R}$ 
define the plane via the Hessian normal form $\mathbf{n}_i^\top \mathbf{p} + d_i = 0$, where $\mathbf{p} \in \mathbb{R}^3$ is any point on the plane and $d_i$ represents the signed distance from the origin}{Here, $\mathbf{n}_i$ is the unit normal of the plane and $d_i$ is the signed perpendicular  distance from the SLAM frame origin (the first keyframe) to the plane}.
\change{While $l_i$ and $t_i$ represent the length and thickness of the wall extracted from the BIM, they are used to compute the centroid $\mathbf{c}_i$ for spatial data association and to derive opposing boundary planes, respectively.}
{The length $l_i$ and thickness $t_i$ are used to compute the centroid $\mathbf{c}_i$ for data association and to derive opposing boundary planes, respectively.}
We denote $\mathbf{W}$ as a set of walls, with $\mathbf{W_{\text{BIM}}}$ referring to BIM walls and $\mathbf{W_S}$ to detected walls, and use $\pi$ to refer specifically to the planar component of a wall in the back-end optimization.
\add{For $\mathbf{W}_{\text{BIM}}$, the wall parameters $(\mathbf{n}_i, d_i, \mathbf{c}_i, l_i, t_i)$ are extracted offline from the BIM, where this geometry is directly encoded.}
Three primary modules comprise the architecture: the SLAM Backbone (Section \ref{sec:vS-Graph}), Wall Matching (Section \ref{sec:wall_matching}), and the BIM-Constrained Back-End (Section \ref{sec:bim_backend}).

\subsection{SLAM Backbone}
\label{sec:vS-Graph}
Our system utilizes \textit{vS-Graphs}~\cite{tourani2025vsgraphs} as a backbone, which provides a feature-based front-end derived from \textit{ORB-SLAM3}~\cite{orb3} and a hierarchical graph-based back-end for structural mapping.
For full details on these base modules, we refer the reader to the original works.
The front-end maintains local consistency by processing RGB-D frames. 
\change{Key components include the \textit{Scene Segmentor} from 
\textit{vS-Graphs}~\cite{tourani2025vsgraphs}, which leverages YOSO~\cite{yoso} 
to extract observed $\mathbf{W_S}$ from depth data, and the \textit{ATLAS} 
map manager from \textit{ORB-SLAM3}~\cite{orb3}, which maintains the evolving 
map and the BIM reference.}
{The \textit{Scene Segmentor} module applies 
panoptic semantic segmentation~\cite{yoso} to identify wall pixels in the 
RGB frame and fits wall planes $\pi_i = (\mathbf{n}_i, d_i)$ via RANSAC on 
the corresponding depth point cloud, producing the observed wall set 
$\mathbf{W_S}$.
\addtwo{Restricting RANSAC to wall-classified points ensures the resulting planes correspond to actual walls, avoiding false wall detections on non-wall planar regions such as cabinets or panels, common in cluttered construction sites}.} 
The back-end employs a factor graph encoding geometric and structural scene information:

\begin{itemize}
    \item \textbf{Keyframes $\mathbf{X}$:} Camera poses as $\mathrm{SE}(3)$ nodes connected via visual odometry and loop-closure constraints.
    \item \textbf{Map points $\mathbf{P}$:} 3D landmarks constrained by reprojection factors linking them to keyframes.
    \item \textbf{Walls $\mathbf{W_S}$:} Structural wall planes extracted by the \textit{Scene Segmentor}. These are mathematically tied to the map points that define their surface, ensuring geometric consistency between landmarks and structural planes.
\end{itemize}

\subsection{Wall Matching}

\label{sec:wall_matching}
The procedure is detailed in Algorithm~\ref{alg:plane_matching}, organized into two stages: (1) initial alignment and (2) continuous matching. First, the module initializes alignment by estimating the transformation \addtwo{$\leftidx{^S}{\hat{T}}_{BIM}$} between the BIM frame and SLAM frame $\mathcal{S}$. All BIM walls $\mathbf{W_{\text{BIM}}}$  are then transformed to $\mathcal{S}$ and stored in the Atlas Manager. Expressing both detected walls $\mathbf{W_S}$ and BIM walls $\mathbf{W_{\text{BIM}}}$  in this common frame enables \addtwo{their} continuous association \addtwo{into the matched set $\mathcal{M}$}.

\begin{algorithm}[h]
\caption{Wall Matching Procedure}
\label{alg:plane_matching}
\begin{algorithmic}[1]
\State \textbf{Input:} $\mathbf{W_{\text{BIM}}}$, $\mathbf{W_S}$, $(id_k, id_l)$
\State \textbf{Output:} $\leftidx{^S}{\hat{T}}_{BIM}$, $\mathcal{M}$

\medskip
\State \textbf{Step 1: Initial Alignment}
\Repeat
   \State Find $w_{S_i}, w_{S_j}$ with $|\theta(\mathbf{n}_{S_i}, \mathbf{n}_{S_j}) - 90^\circ| < \tau_\perp$
    \State Estimate $\leftidx{^S}{\hat{T}}_{BIM}$ via Least Square Method
\Until{residual $< \tau_{init}$}

\medskip
\State \textbf{Step 2: Continuous Matching}
\For{each $w_S \in \mathbf{W_S}$ }
    \For{each $w_{\text{BIM}} \in \mathbf{W_{\text{BIM}}}$ satisfying: \State $d_\text{plane} \le \tau_p$ \& $d_\text{lat} \le \tau_c$}
    \State Compute combined score: $s(w_S,w_{\text{BIM}})$
    \State Best result: $w_{\text{BIM}}^\ast = \arg\min s(w_S,w_{\text{BIM}})$
    \State $\mathcal{M} \leftarrow (w_S, w_{\text{BIM}}^\ast)$
    \EndFor  
\EndFor
\end{algorithmic}
\end{algorithm}

\subsubsection*{\textbf{Initial Alignment}}
\addtwo{The system requires a human operator to specify, via the BIM IDs $id_k$ and $id_l$, the two perpendicular BIM walls corresponding to the first two walls the system will observe. This is a design choice that allows site personnel using the BIM to identify these walls directly from the model, enabling immediate operation at construction sites without extensive prior mapping. The system then continues running the SLAM backbone until two detected walls $(w_{S_i}, w_{S_j})$ whose supporting planes are nearly perpendicular (within $\tau_\perp$, Table~\ref{tab:parameters}) are found and paired with the two specified BIM walls.}
Two perpendicular walls provide a linearly independent system sufficient to estimate the transformation $\leftidx{^S}{\hat{T}}_{BIM} \in \mathrm{SE}(3)$ between coordinate frames by minimizing residuals between corresponding planes via least squares:
\begin{equation}
\leftidx{^S}{\hat{T}}_{BIM} = \arg\min_{T \in \mathrm{SE}(3)} 
\sum_{k=1}^{2} \big\| f(T, \pi_{BIM_k}) - \pi_{S_k} \big\|^2
\end{equation}
\noindent where $f(T,\pi_{BIM})$ applies the transform $T=(R,\mathbf{t})$ to the BIM plane $\pi_{BIM}$ . After $\leftidx{^S}{\hat{T}}_{BIM}$ is estimated, applying it to $\mathbf{W_{\text{BIM}}}$ yields the transformed set $\mathbf{\leftidx{^S}{{W}}_{BIM}}$.
The transformation is accepted once its residual falls below $\tau_{init}$ (Table~\ref{tab:parameters}). Since BIM walls are vertical, their plane equations constrain rotation and translation primarily within the horizontal plane; the vertical component is instead assumed fixed at system setup, aligning the BIM and SLAM frames.

\add{The role of this initial transformation is to bring the \textit{as-planned} and \textit{as-built} models close enough to enable subsequent wall matches. These matches, introduced as wall-to-wall factors (Section~\ref{sec:bim_backend}), continuously refine the alignment between the two models throughout the trajectory.}

\subsubsection*{\textbf{Continuous Matching}}
After initialization, the system updates wall associations as new walls are detected, refining the alignment between the SLAM map and the BIM. For each detected wall $w_{S_i}$, two complementary measures are evaluated against each candidate BIM wall $w_{BIM_j}$:

\textbf{1) Plane-parameter distance (PPD).}  
The distance between a detected plane $\pi_{S_i}$ and a BIM plane $\pi_{BIM_j}$ is computed using their minimal parameter vectors as:
\begin{equation}
\label{eq:ppd}
d_\text{plane}(\pi_{S_i},\pi_{BIM_j}) = \sqrt{(\pi_{S_i} \ominus \pi_{BIM_j})^\top (\pi_{S_i} \ominus \pi_{BIM_j})} 
\end{equation}
\change{where $\ominus$ computes the minimal distance between plane parameters, accounting for sign ambiguity in the Hessian normal form. This metric evaluates infinite surface alignment.}
{where $\ominus$ is the plane difference operator: $\pi_a \ominus \pi_b = (\phi_b, \theta_b, d_a - d_b)^\top$, with $\phi_b, \theta_b$ the azimuth and elevation of $R_a^\top \mathbf{n}_b$, and $R_a = R_z(\phi_a) R_y(-\theta_a)$ aligning the $x$-axis with $\mathbf{n}_a$ via its azimuth $\phi_a$ and elevation $\theta_a$. This metric combines the angular and distance residuals between the planes.}

\textbf{2) Lateral Centroid Distance (LCD).}
\change{Plane-based similarity alone is insufficient for data association, since accumulated drift can cause the infinite plane of a detected wall to align with an incorrect BIM wall located in a different room.}{
Plane-based similarity alone is insufficient for data association: in environments with repetitive layouts, walls in different rooms often share similar plane equations, and accumulated drift can further cause the infinite plane of a detected wall to align with an incorrect BIM wall located in a different room. 
To disambiguate walls with similar plane parameters, we enforce spatial consistency of the finite wall segments by measuring the tangential displacement between their centroids.}

Specifically, we define the lateral centroid distance $d_\text{lat}$ as the Euclidean distance between centroids projected onto the supporting plane of the detected wall:
\begin{equation}
d_{\mathrm{lat}}(w_{S_i}, w_{\text{BIM}_j}) =
\sqrt{
\|\Delta\mathbf{c}\|^2 -
\left(\Delta\mathbf{c}^\top \mathbf{n}_{S_i}\right)^2
},
\end{equation}
This metric favors the closest architectural counterpart over lateral offsets, even for similar plane equations.

\textbf{3) Combined score.}  
PPD and LCD are combined into a normalized weighted score $s(w_{S_i},w_{BIM_j})$ to determine the best match:
\begin{equation}
s(w_{S_i}, w_{\text{BIM}_j}) = \alpha 
\frac{d_\text{plane}(\pi_{S_i}, \pi_{\text{BIM}_j})}{\tau_p} + 
(1-\alpha) \frac{d_{\text{lat}}(w_{S_i}, w_{\text{BIM}_j})}{\tau_c}
\end{equation}
where $\tau_p$ and $\tau_c$ are normalization constants that ensure the two metrics are comparable, and $\alpha \in [0,1]$ controls the relative importance of surface alignment versus spatial proximity. The score is normalized such that $s = 0$ indicates perfect alignment ($d_\text{plane} = 0$, $d_\text{lat} = 0$), and $s = 1$ represents the maximum acceptable deviation.
To ensure data association robustness, only pairs satisfying $d_\text{plane} < \tau_p$ and $d_\text{lat} < \tau_c$ are considered. The candidate $w_{BIM_j}$ with the lowest score $s(w_{S_i},w_{BIM_j})$ is selected, and all associations are collected into the set $\mathcal{M} = \{ (w_{S_i}, w_{BIM_j}) \}$.
\add{Once a detected wall is associated with a BIM wall, the match is fixed for the remainder of the sequence and is not re-evaluated in subsequent frames. The matching procedure therefore fires only when a new wall is detected.}

\subsection{BIM-Constrained Back-End Integration}
\label{sec:bim_backend}
\change{This module incorporates matched walls into the factor graph 
as additional nodes. By establishing edges between corresponding wall 
pairs, the system constrains the evolving map to the 
\textit{as-planned} layout.}{This module corrects drift in the \textit{as-built} map by constraining it to the \textit{as-planned} layout.}

\subsubsection*{\textbf{Graph Construction}}
\change{The back-end factor graph integrates keyframes $\mathbf{X}$, 
map points $\mathbf{P}$, and detected structural walls $\mathbf{W_S}$. 
After the initial alignment (Section~\ref{sec:wall_matching}), BIM 
walls $\mathbf{W_{\text{BIM}}}$ are introduced as fixed wall nodes in 
the graph as they represent the trusted \textit{as-planned} reference, 
while detected walls $\mathbf{W_S}$ remain optimizable.}
{After the initial alignment (Section~\ref{sec:wall_matching}), ivS-Graphs constructs a global graph that integrates the matched wall pairs into the back-end optimization. 
The graph contains two types of wall nodes: BIM walls $\mathbf{W_{\text{BIM}}}$, held fixed as the \textit{as-planned} reference, and their matched detected walls $\mathbf{W_S}$, which are jointly optimized with the keyframes $\mathbf{X}$. Edges constrain each matched pair $(w_{S_i}, w_{\text{BIM}_j}) \in \mathcal{M}$ as well as the detected walls to their associated keyframes.}

Wall-to-wall factors are created from the associations $\mathcal{M}$ found during matching. The cost function is:
\begin{equation}
c_{\text{wall-wall}}(w_{S_i},w_{BIM_j}) = \big\| w_{S_i} \ominus w_{BIM_j} \big\|^2_{\mathrm{\Lambda}_{w_{ij}}}
\label{eq:plane_to_plane_cost}
\end{equation}
\add{where $\ominus$ is the plane difference operator defined after Equation~\ref{eq:ppd}}
and $\mathrm{\Lambda}_{w_{ij}}$ is the \addtwo{information matrix} associated with the uncertainty of the plane association. This uncertainty is derived from the wall matching score $s(w_{S_i},w_{BIM_j})$ defined in Section~\ref{sec:wall_matching}. Since lower matching scores indicate better alignment (where $s = 0$ represents perfect match), $\mathrm{\Lambda}_{w_{ij}}$ \addtwo{grows inversely with $s$ to assign larger information to high-confidence matches and smaller information to uncertain ones}:
\begin{equation}
\mathrm{\Lambda}_{w_{ij}} = \addtwo{\dfrac{\mathrm{I} \cdot \beta}{\max(s(w_{S_i}, w_{BIM_j}),\ s_{\min})}}
\end{equation}
where $\mathbf{I}$ is the identity matrix and $\beta$ is a global weighting parameter that balances architectural constraints against visual odometry\addtwo{, and $s_{\min} = 10^{-3}$ is a floor for numerical stability, bounding the maximum information as $s \to 0$ for a near-perfect match}. This formulation ensures well-aligned walls (low $s$, large $\mathrm{\Lambda}_{w_{ij}}$) exert stronger constraints, while ambiguous associations (high $s$, small $\mathrm{\Lambda}_{w_{ij}}$) are automatically down-weighted in the optimization.

\subsubsection*{\textbf{Graph Optimization}}
The factor graph is periodically optimized to maintain consistency and incorporate BIM wall constraints as new associations become available.  

The total cost function of the system is composed of three types of residuals: visual reprojection errors, pose-to-wall constraints, and BIM wall-to-wall factors (our contribution):
\begin{align}
c_{\text{total}} = \sum_{x,p} c_{\text{reproj}}&(x,p) 
+ \sum_{x,w_{S}} \big\| (x \cdot w_{S}) \ominus \tilde{w}_{S} 
\big\|^2_{\Lambda_{\tilde{w}}} \nonumber \\
&+ \sum_{(w_{S},w_{BIM}) \in \mathcal{M}} 
\big\| w_{S} \ominus w_{BIM} \big\|^2_{\Lambda_{w}}
\end{align}
\add{where the second term constrains each keyframe pose $x \in SE(3)$ 
by transforming the globally estimated wall $w_S$ into the keyframe 
frame and comparing it with the locally observed wall $\tilde{w}_S$, 
with $\ominus$ defined in Equation~\ref{eq:ppd}.} The optimizer adjusts 
the variables $\{\mathbf{X},\mathbf{P},\mathbf{W_S}\}$ to minimize 
$c_{\text{total}}$. The gradient of each residual propagates corrections 
to the connected keyframes and map points, ensuring that matched BIM 
walls influence the SLAM map according to their confidence. 
\add{After optimization, keyframe poses and map points are updated in the map, while BIM walls remain fixed. The wall-to-wall factors therefore correct keyframe poses globally, refining the alignment between the \textit{as-planned} and \textit{as-built} models throughout the trajectory and bounding the drift accumulated during operation}. 

Associations with large residuals are automatically down-weighted using a robust Huber kernel~\cite{GALLEGO2021174}, which reduces their influence on the optimization.  
This is particularly important in construction environments, where partial structures or occlusions can cause incorrect wall detections and, consequently, erroneous wall associations.

\section{Experimental Results}
\label{sec_evaluation}

\subsection{Validation Methodology}
\label{sec_eval_setup}

\begin{table}[b]
    \centering
    \caption{Summary of evaluation sequences. 
    \#W and \#S denote the number of visible walls and distinct navigable spaces (rooms and corridors), respectively. Area, Length ($L$), and Duration ($D$) are in $m^2$, $m$, and $s$.
    CS1--CS3-Furn are distinct construction sites at different completion stages; office1 and office2 are separate buildings with multiple sequences each. }
    \begin{tabular}{l|l|c|c|c|c|c}
        \toprule
            \textbf{Env.} & \textbf{Sequence} & \textbf{\#W} & \textbf{\#S} & \textbf{Area ($\mathbf{m}^2$)} & \textbf{L (m)} & \textbf{D(s)} \\
        \midrule
        \multirow{8}{*}{\shortstack{\textbf{Offices}}} 
            & office1-1 & 10 & 4 & 100 & 90 & 670 \\
            & office1-2 & 12 & 3 & 109 & 48 & 265 \\
            & office1-3 & 12 & 4 & 113 & 99 & 524 \\
            & office1-4 & 6 & 2 & 54 & 37 & 169 \\
        \cmidrule(lr){2-7}
            & office2-1 & 10 & 3 & 168 & 68 & 268 \\
            & office2-2 & 10 & 3 & 168 & 73 & 370 \\
            & office2-3 & 10 & 3 & 70 & 41 & 172 \\
            & office2-4 & 4 & 4 & 40 & 53 & 260 \\
        \midrule
        \multirow{3}{*}{\shortstack{\textbf{Constr.}\\\textbf{Sites}}} 
            & CS1       & 16 & 5 & 48 & 51 & 377 \\
        \cmidrule(lr){2-7}
            & CS2-Mid   & 17 & 5 & 68 & 46 & 215 \\
        \cmidrule(lr){2-7}
            & CS3-Furn  & 13 & 4 & 95 & 30 & 180 \\
        \midrule
            \multicolumn{2}{c|}{\textbf{Total}} & 120 & 40 & 1033 & 642 & 3470 \\
        \bottomrule
    \end{tabular}
    \label{tbl_dataset}
\end{table}

\textbf{Datasets.}
Public indoor datasets (e.g., \cite{dataset_scannet}) typically provide RGB-D data, but to the best of our knowledge, no public dataset includes both RGB-D and the corresponding BIM, which limits the evaluation of our system. 
\add{To address this, we collected real-world sequences with the SMapper device~\cite{smapper}, a portable acquisition platform integrating an \textit{Intel RealSense D435i} RGB-D camera and an \textit{Ouster OS0-64} 3D LiDAR. ivS-Graphs uses only the RGB-D stream as input; the LiDAR point clouds are used for ground-truth trajectory generation (Section~IV-A). This type of camera is commonly mounted on robotic and handheld platforms.}

The sequences span multiple indoor layouts as summarized in Table~\ref{tbl_dataset}, including two office buildings with varying trajectories and three construction sites at different completion stages, \add{all within indoor structured environments where architectural walls are already in place}: early phase with exposed installations (\textit{CS1}), mid-phase with ongoing work (\textit{CS2-Mid}), and near-finished with furniture present (\textit{CS3-Furn}).

\begin{table}[t]
\centering
 \vspace{0.25cm}
\caption{Key Parameters and Thresholds for ivS-Graphs}
\label{tab:parameters}
\begin{tabular}{@{}lllc@{}}
\toprule
\textbf{Param.} & \textbf{Value} & \textbf{Description} & \textbf{Section} \\ \midrule
$\tau_{\perp}$ & 0.17 rad & Perpendicularity threshold for initialization & \ref{sec:wall_matching} \\
$\tau_{init}$ & 0.30 m & Max. residual for initial alignment & \ref{sec:wall_matching} \\
$\tau_p$ & 0.77 m & Plane-parameter distance threshold & \ref{sec:wall_matching} \\
$\tau_c$ & 6.00 m & Lateral centroid distance threshold & \ref{sec:wall_matching} \\
$\alpha$ & 0.70 & Weighting for plane vs. spatial proximity & \ref{sec:wall_matching} \\
$\beta$ & $10^{-2}$ & \addtwo{Base weight for wall-to-wall factors} & \ref{sec:bim_backend} \\ \bottomrule
\end{tabular}
\end{table}

\textbf{Baselines.}
\change{Since no existing visual SLAM system integrates BIM priors, we evaluate against established RGB-D SLAM baselines: ORB-SLAM3~\cite{orb3}, BAD-SLAM~\cite{badslam}, and DROID-SLAM~\cite{droidslam} are selected for their reported performance and code availability.}{Since no existing visual SLAM system integrates BIM priors, we evaluate against representative RGB-D visual SLAM methods that do not rely on external architectural priors. The selected baselines cover the main RGB-D visual SLAM paradigms: feature-based (\textit{ORB-SLAM3}~\cite{orb3}), dense direct (\textit{BAD-SLAM}~\cite{badslam}), and learning-based (\textit{DROID-SLAM}~\cite{droidslam}).}
\add{Additionally, the comparison with \textit{vS-Graphs}~\cite{tourani2025vsgraphs}, one of the best-performing structure-aware visual SLAM systems that integrates wall planes into the factor graph, allows us to directly measure the effect of adding BIM constraints as our system builds on it.}

\textbf{Implementation Details.} All algorithms were executed on a workstation with an \textit{Intel Core i9-11950H} (2.60\,GHz), an \textit{NVIDIA T600 Mobile GPU}, and 32\,GB RAM. 
\add{The BIM structural data is extracted from an IFC model.
For Revit models, an automated Dynamo script converts them into IFC.
Only architectural walls are extracted; other BIM elements such as doors and windows are not used.}

\textbf{System parameters}. The parameters in Table~\ref{tab:parameters} were tuned empirically based on physical constraints: $\tau_p$ and $\tau_c$ reflect both the expected RGB-D sensor uncertainty and typical dimensions of architectural wall segments. The weighting factor $\alpha$ prioritizes geometric alignment (PPD) over spatial proximity (LCD), while $\beta$ balances BIM constraints against visual odometry, ensuring that well-matched walls exert meaningful corrections without overriding visual observations. All parameters are kept constant across every experiment and environment, demonstrating the generalization capability of the proposed configuration.

\textbf{Trajectory Estimation Performance.}
We evaluate trajectory accuracy using the Absolute Trajectory Error (ATE), as drift accumulation directly degrades reconstruction quality.
As no external input of ground truth was available, reference trajectories are generated by processing Ouster OS0-64 LiDAR scans with S-Graphs~\cite{bavle2025s}, which achieves an average ATE of $4.16\,\text{cm}$ in similar indoor environments. This is substantially more accurate than any visual method evaluated.

\add{\textbf{Wall-Matching Precision.} We report the wall-matching precision per sequence to assess association correctness, defined as the percentage of detected walls whose predicted BIM match agrees with a manually annotated ground-truth match.}

\begin{table}[t!]
  \vspace{0.25cm}
    \centering
    \footnotesize 
    \setlength{\tabcolsep}{2.2pt} 
    \caption{\acf{ATE} in meters ($\mathrm{m}$) for the evaluated 
    \acs{VSLAM} algorithms. The best and second-best results are 
    \textbf{boldfaced} and \underline{underlined}, respectively. 
    Values are averaged over five runs. The minimum completion rate 
    of each sequence is set to 50\%, and dashes indicate unavailable 
    data due to tracking failure. The \textit{Diff.} column reports 
    the improvement of \textit{ivS-Graphs} over \textit{vS-Graphs} 
    baseline\add{, and final column Prec. (\%) reports the wall-matching precision}.}
    \resizebox{\columnwidth}{!}{
    \begin{tabular}{l|ccccc|c||c}
        \toprule
            \multirow{4}{*}{{\textbf{Sequence}}} & 
            \multicolumn{6}{c|}{\textbf{ATE ($\mathrm{m}$)}} &
            \multirow{4}{*}{\shortstack{\textbf{Prec.}\\(\%)\\(ours)}} \\
        \cmidrule{2-7}
            & {\shortstack{\textit{BAD} \\ \textit{SLAM} \\ 
            \cite{badslam}}} 
            & {\shortstack{\textit{ORB-} \\ \textit{SLAM3} \\ 
            \cite{orb3}}} 
            & {\shortstack{\textit{DROID-} \\ \textit{SLAM} \\ 
            \cite{droidslam}}}
            & {\shortstack{\textit{\vgraphs} \\ 
            \cite{tourani2025vsgraphs}}} 
            & {\shortstack{\textit{\ivgraphs}\\(ours)}} 
            & \textit{Diff. (\%)} & \\
        \midrule
           office1-1 & -     & 0.462 & 0.541 & \underline{0.343} & \textbf{0.226} & {+33.99\%} & 90 \\
           office1-2 & 1.653 & 0.365 & 0.533 & \underline{0.353} & \textbf{0.183} & {+48.15\%} & 100 \\
           office1-3 & -     & 0.272 & 0.360 & \underline{0.253} & \textbf{0.175} & {+30.83\%} & 91 \\
           office1-4 & -     & \underline{0.101} & 0.109 & \textbf{0.090} & 0.120 & {-33.33\%} & 100 \\
         \midrule
            office2-1 & 1.879 & \textbf{0.160} & 0.395 & 0.223 & \underline{0.214} & {+4.04\%} & 100 \\
            office2-2 & -     & 0.461 & 0.634 & \underline{0.348} & \textbf{0.224} & {+35.76\%} & 90 \\
            office2-3 & -     & 0.393 & \textbf{0.158} & 0.280 & \underline{0.176} & {+36.85\%} & 90 \\
            office2-4 & 7.926 & 0.553 & \underline{0.192} & 0.386 & \textbf{0.123} & {+68.10\%} & 100 \\
        \midrule
            CS1       & 2.798 & 0.136 & 0.179 & \underline{0.112} & \textbf{0.100} & {+10.51\%} & 92 \\
        \midrule
            CS2-Mid   & 1.383 & 0.213 & 0.494 & \underline{0.158} & \textbf{0.156} & {+1.22\%} & 100 \\
        \midrule
            CS3-Furn  & -     & 0.241 & \underline{0.205} & 0.254 & \textbf{0.149} & {+41.38\%} & 78 \\
        \midrule
             \textbf{Mean} & - & 0.305 & 0.346 & \underline{0.254} & \textbf{0.168} & {+25.23\%} & \textbf{92.8} \\
        \bottomrule
    \end{tabular}
    }
    \label{tbl_eval}
\end{table}

\textbf{Mapping Performance.}  
We evaluate the quality of the reconstructed maps by computing the Root Mean Square Error (RMSE) between the reconstructed point cloud and the reference BIM point cloud, which serves as ground truth.  
This metric directly reflects the geometric consistency of the map with respect to the \textit{as-planned} design.

\textbf{Robustness to Model Incompleteness and Deviations.}
To evaluate the system's resilience to discrepancies between the \textit{as-planned} BIM and the \textit{as-built} site, we simulate two scenarios: model incompleteness and plan deviations. First, we simulate partially built environments by omitting 20\% and 30\% of the BIM walls from the matching process. Second, we perturb all BIM walls\addtwo{, including the two used for initial alignment, to evaluate the system’s tolerance to \textit{as-built} vs. \textit{as-planned} discrepancies and the robustness of the initial alignment}. For each wall, the specific deviation is sampled from a uniform distribution, \changetwo{$[0, \Delta d]$ and $[0, \Delta \theta]$}{$U[-\Delta d, +\Delta d]$ for position and $U[-\Delta \theta, +\Delta \theta]$ for orientation}. \addtwo{To avoid dependence on a single perturbation sample, we repeat each sequence and condition on five deviated BIM models, each generated by re-sampling the per-wall deviations.
}

\textbf{Runtime Analysis.}
To verify real-time operation, we measured the processing frame rate (FPS) and the time required for initial alignment across four representative sequences.

\subsection{Results and Discussion}
\label{sec_results_discussion}

\subsubsection{\textbf{Trajectory Estimation Performance}}
\begin{figure}[t]
    \centering
    \includegraphics[width=0.4\textwidth]{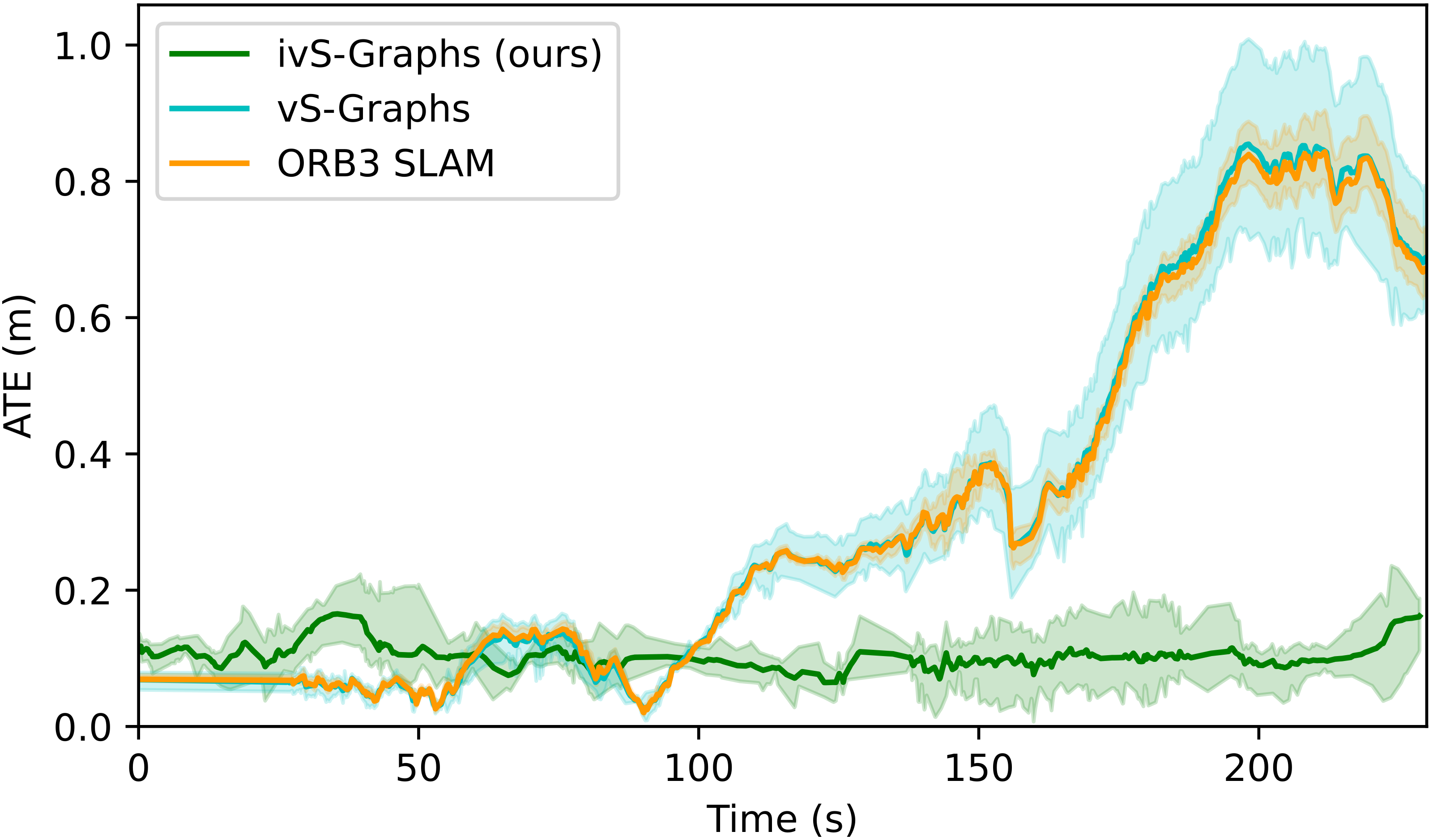}
    \caption{Accumulated trajectory drift error over time in \textit{office1-2} sequence, showing the ATE for baselines and \textit{ivS-Graphs}. 
    }
    \label{fig:accumulated_drift}
\end{figure}

Table~\ref{tbl_eval} summarizes the ATE results. \textit{ivS-Graphs} reduces mean ATE by 25.23\% over \textit{vS-Graphs} (second-best), with improvements in 10 of 11 sequences. Improvement scales with trajectory length: sequences exceeding $50\,\text{m}$ (e.g.\ \textit{office1-1}) show average improvement of 30.54\%, while shorter sequences show 18.85\%, confirming that BIM constraints primarily combat drift accumulation in extended operation. This drift-bounding effect is visible in Fig.~\ref{fig:accumulated_drift}: the baseline error grows linearly to $0.8\,\text{m}$, while \textit{ivS-Graphs} stays below $0.2\,\text{m}$ through continuous wall-matching constraints. The single degraded sequence (\textit{office1-4}, -33.33\%) is also one of the shortest ($37.69\,\text{m}$), where minimal drift gives BIM constraints little to correct.

\begin{figure}[b!]
    \centering
    \includegraphics[width=\linewidth]{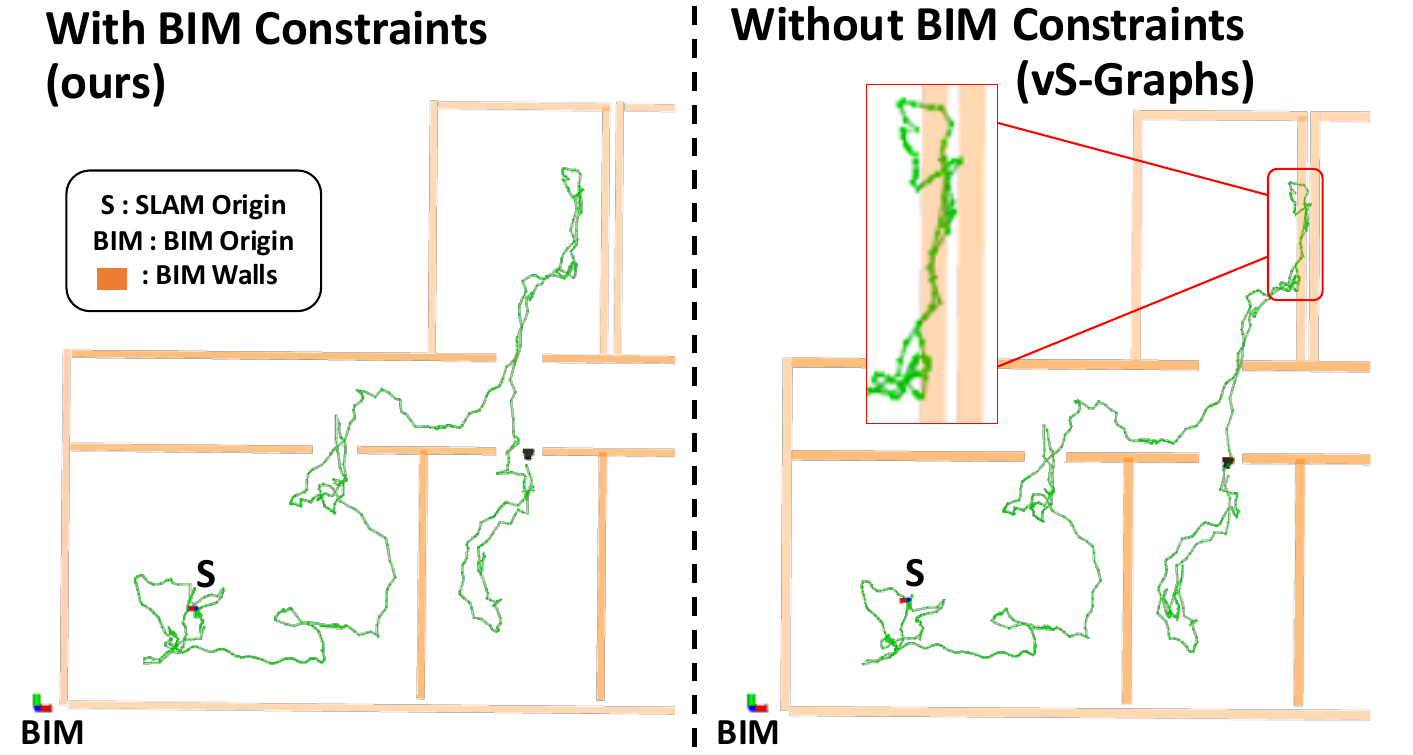}
    \caption{Top-down overlay trajectory overlay on \textit{office1-1}. Left: \textit{ivS-Graphs} (with BIM). Right: \textit{vS-Graphs} (init. align. only)}
    \label{fig:qualitative_trajectory}
\end{figure}

The qualitative comparison in Fig.~\ref{fig:qualitative_trajectory} further confirms this trend, showing a top-down overlay of the estimated trajectories on the \textit{office1-1} sequence. For the \textit{vS-Graphs} baseline (right), only the initial semantic alignment is used to display the \textit{as-planned} data. The trajectory estimated by \textit{ivS-Graphs} (left) is topologically consistent with the floor layout, whereas \textit{vS-Graphs} exhibits deviations, with the trajectory crossing through BIM walls. 
The remaining baselines also struggled: BAD-SLAM~\cite{badslam} failed to track in 6 out of 11 sequences due to textureless surfaces, while DROID-SLAM~\cite{droidslam} remained more robust but still exhibited significant drift.

\add{\subsubsection{\textbf{Wall-Matching Precision}}
Table~\ref{tbl_eval} reports the wall-matching precision per sequence, achieving a mean of 92.8\%. Most mismatches occur between coplanar walls from adjacent rooms. Since the back-end optimization uses only the planar component $\pi = (\mathbf{n}, d)$ of each wall, coplanar walls produce identical constraints and the mismatch does not affect trajectory accuracy. The single true mismatch in \textit{office2-3} is successfully down-weighted by the Huber kernel, as confirmed by its competitive ATE of $0.176\,\text{m}$.}

\subsubsection{\textbf{Mapping Performance}}  
Table~\ref{tbl_RMSE} summarizes the mapping accuracy results. 
\textit{ivS-Graphs} achieves lower point cloud RMSE than the \textit{vS-Graphs} backbone in 9 out of 11 sequences, with improvements on those sequences ranging from +0.99\% to +17.88\%.
The gains follow a clear size-dependent pattern: in larger environments such as \textit{office1-1} (+17.88\%) and \textit{CS1} (+9.63\%), where trajectory drift accumulates over extended operation, BIM constraints produce consistent and meaningful map improvements. In smaller environments, where the baseline trajectory is already accurate, the contribution of BIM integration is marginal. This also explains why the relative gains are smaller than those observed for ATE: BIM constraints reduce drift-induced map distortion but cannot compensate for the per-frame depth noise of the RGB-D sensor, which remains an independent source of map error.

\begin{table}[t!]
 \vspace{0.25cm}
    \centering
    \footnotesize
    \setlength{\tabcolsep}{2.2pt}
    \caption{Root Mean Square Error (RMSE) in meters of reconstructed point clouds for the evaluated \acs{VSLAM} algorithms. The best and second-best results are \textbf{boldfaced} and \underline{underlined}, respectively. Values are averaged over 5 runs. The minimum completion rate of each sequence is set to 50\%, and dashes indicate missing data due to tracking failure. The final column reports the improvement of \textit{ivS-Graphs} over \textit{vS-Graphs}.
    }
    
    \resizebox{\columnwidth}{!}{
    \begin{tabular}{l|ccccc|c}
        \toprule
            \multirow{4}{*}{{\textbf{Sequence}}} & \multicolumn{6}{c}{\textbf{Point Cloud RMSE ($\mathrm{m}$)}} \\
        \cmidrule{2-7}
            & {\shortstack{\textit{BAD} \\ \textit{SLAM} \\ \cite{badslam}}} 
            & {\shortstack{\textit{ORB-} \\ \textit{SLAM 3} \\ \cite{orb3}}} 
            & {\shortstack{\textit{DROID-} \\ \textit{SLAM} \\ \cite{droidslam}}} 
            & {\shortstack{\textit{\vgraphs} \\ \cite{tourani2025vsgraphs}}} 
            & {\shortstack{\textit{\ivgraphs}\\(ours)}} 
            & \textit{Diff. (\%)} \\
        \midrule
           office1-1  & -     & 0.416 & 0.430 & \underline{0.403} & \textbf{0.331} & {+17.88\%} \\
           office1-2  & 0.442 & 0.537 & \underline{0.435} & 0.444 & \textbf{0.396} & {+10.72\%} \\
           office1-3  & -     & 0.360 & 0.395 & \textbf{0.340} & \underline{0.354} & {-4.15\%} \\
           office1-4  & -     & 0.349 & \underline{0.323} & 0.327 & \textbf{0.306} & {+6.35\%} \\
         \midrule
            office2-1 & 0.394 & 0.407 & \textbf{0.371} & 0.432 & \underline{0.382} & {+11.55\%} \\
            office2-2 & -     & 0.487 & \textbf{0.404} & \underline{0.470} & 0.473 & {-0.80\%} \\
            office2-3 & -     & \underline{0.412} & 0.466 & 0.481 & \textbf{0.411} & {+14.60\%} \\
            office2-4 & -     & \textbf{0.386} & \underline{0.460} & 0.546 & 0.504 & {+7.64\%} \\
        \midrule
            CS1       & 0.299 & 0.282 & \textbf{0.252} & 0.307 & \underline{0.277} & {+9.63\%} \\
        \midrule
            CS2-Mid & \textbf{0.357} & 0.425 & 0.268 & 0.378 & \underline{0.374} & {+0.99\%} \\
        \midrule
            CS3-Furn  & -     & 0.588 & \underline{0.296} & 0.298 & \textbf{0.287} & {+3.62\%} \\
        \midrule
             \textbf{Mean} & - & 0.422 & \underline{0.373} & 0.402 & \textbf{0.371} & \multicolumn{1}{c}{{+7.14\%}} \\
        \bottomrule
    \end{tabular}
    }
    \label{tbl_RMSE}
\end{table}

\begin{table}[b]
    \centering
    \caption{ATE [m] under varying BIM completeness. Nominal uses the full model; 20\% and 30\% Miss. omit walls. Rel. Err. shows ATE increase from Nominal to 30\% Missing case.}
    \label{tab:ate_missing_walls}
    \begin{tabular}{l|c|c|c|c}
    \toprule
    \textbf{Sequence} & \textbf{Nominal} & \textbf{20\% Miss.} & \textbf{30\% Miss.} & \textbf{Rel. Err. (\%)} \\
    \midrule
    office1-1 & 0.226 & 0.244 & 0.253 & +11.9\% \\
    CS1 & 0.100 & 0.104 & 0.094 & -6.0\% \\
    CS2-Mid & 0.156 & 0.188 & 0.166 & +6.4\% \\
    CS3-Furn & 0.149 & 0.157 & 0.162 & +8.7\% \\
    \midrule
    \textbf{Mean} & \textbf{0.158} & \textbf{0.173} & \textbf{0.169} & \textbf{+5.3\%} \\
    \bottomrule
    \end{tabular}
\end{table}

\subsubsection{\textbf{Robustness to Model Incompleteness and Deviations}} 
Table~\ref{tab:ate_missing_walls} demonstrates the ATE results for robustness in partially constructed environments. With 30\% of walls missing, the system experiences only 5.3\% average error increase. This minimal degradation occurs because missing walls reduce structural constraints without introducing erroneous information; the remaining 70\% continue to provide drift correction.

\begin{table}[t]
    \centering
     \vspace{0.25cm}
     \caption{\addtwo{ATE [m] and Match Rate (MR \%) under geometric deviations, reported as ATE / MR. Moderate ($\pm 0.2\,\text{m}$, $\pm 5^\circ$) and Severe ($\pm 0.8\,\text{m}$, $\pm 15^\circ$) perturbations, averaged over 5 randomized runs ($\pm$ standard deviation), are applied to the BIM walls. Rel. Err. indicates ATE \% increase over Nominal, whose MR is 100\% for all sequences.}}
    \label{tab:deviations}
    \footnotesize
    \resizebox{\columnwidth}{!}{%
    \addtwo{%
    \begin{tabular}{l | c | c | c | cc}
    \toprule
    \multirow{2}{*}{\textbf{Seq.}} & \textbf{Nom.} & \textbf{Moderate} & \textbf{Severe} & \multicolumn{2}{c}{\textbf{Rel. Err. (\%)}} \\
    & ATE & ATE / MR & ATE / MR & \textbf{Mod.} & \textbf{Sev.} \\
    \midrule
    off1-1      & 0.23 & 0.29\,{\scriptsize$\pm$0.05}\,/\,96.0\,{\scriptsize$\pm$5.5} & 0.30\,{\scriptsize$\pm$0.16}\,/\,78.0\,{\scriptsize$\pm$8.4} & +28.3 & +33.4 \\
    CS1         & 0.10 & 0.16\,{\scriptsize$\pm$0.06}\,/\,91.3\,{\scriptsize$\pm$9.5} & 0.16\,{\scriptsize$\pm$0.04}\,/\,85.0\,{\scriptsize$\pm$12.2} & +63.4 & +64.1 \\
    CS2-Mid     & 0.16 & 0.18\,{\scriptsize$\pm$0.04}\,/\,80.0\,{\scriptsize$\pm$13.9} & 0.22\,{\scriptsize$\pm$0.09}\,/\,73.3\,{\scriptsize$\pm$22.4} & +12.5 & +38.0 \\
    CS3-Furn    & 0.15 & 0.18\,{\scriptsize$\pm$0.08}\,/\,94.5\,{\scriptsize$\pm$5.0} & 0.40\,{\scriptsize$\pm$0.26}\,/\,81.8\,{\scriptsize$\pm$6.4} & +20.2 & +167.7 \\
    \midrule
    \textbf{Mean} & \textbf{0.16} & \textbf{0.20 / 90.4} & \textbf{0.27 / 79.5} & \textbf{+28.0} & \textbf{+71.1} \\
    \bottomrule
    \end{tabular}%
    }%
    }
\end{table}

Table~\ref{tab:deviations} evaluates the system's robustness by quantifying the impact of geometric discrepancies on localization accuracy (ATE) and wall association (MR). \addtwo{In every run, initialization succeeded within $\tau_{init}$, keeping the \textit{as-planned} and \textit{as-built} models close enough for the subsequent wall matching to progressively reduce their misalignment.} In the Moderate scenario, which applies realistic construction tolerances to all walls simultaneously, the Match Rate remains high (\changetwo{88.0}{90.4}\%) because deviations fall within association thresholds as they can be misinterpreted as accumulated drift. The back-end optimization incorporates these deviated constraints, resulting in \changetwo{25.8}{28.0}\% average error increase, with the Huber kernel down-weighting the largest residuals preventing catastrophic failure.

The Severe scenario reveals two distinct failure modes. When MR remains high (e.g., CS3-Furn: \changetwo{90.9}{81.8}\% MR), highly deviated walls continue to match and mislead the optimization, causing \changetwo{143.6}{167.7}\% error increase. Conversely, when large deviations trigger rejection (e.g., \changetwo{office1-1: 64.7}{CS2-Mid: 73.3}\% MR), the system relies more on visual baseline, with a lower \changetwo{40.7}{38.0}\% error increase.
This demonstrates that rejecting ambiguous associations outperforms accepting incorrect matches, and that even under deviations far exceeding typical construction tolerances, the system degrades predictably.

\begin{table}[b]
    \centering
    \caption{Runtime performance. FPS is the average processing rate over the trajectory. Align. is the one-time initial registration cost. Match and Opt. (ms) are the mean per-call costs of continuous wall matching (per detected wall) and BIM back-end optimization.}
    \label{tab:runtime}
    \setlength{\tabcolsep}{3pt}
    \begin{tabular}{l|c|ccccc}
    \toprule
    \multirow{2}{*}{\textbf{Sequence}} & \textbf{vS-Graphs} & \multicolumn{5}{c}{\textbf{ivS-Graphs (Ours)}} \\
    \cmidrule(lr){3-7}
    & \textbf{FPS [6]} & \textbf{FPS} & \shortstack{\textbf{Align.}\\\textbf{(ms)}} & \shortstack{\textbf{Match}\\\textbf{(ms)}} & \shortstack{\textbf{Opt.}\\\textbf{(ms)}} & \shortstack{\textbf{Opt.}\\\textbf{calls}} \\
    \midrule
    office1-1   & 23.8 & 21.8 & 0.24 & 0.06 & 843 & 67 \\
    CS1         & 23.3 & 22.5 & 0.23 & 0.07 & 542 & 38 \\
    CS2-Mid     & 24.2 & 23.9 & 0.34 & 0.06 & 321 & 22 \\
    CS3-Furn    & 25.5 & 24.9 & 0.15 & 0.05 & 442 & 18 \\
    \midrule
    \textbf{Mean} & \textbf{24.2} & \textbf{23.3} & \textbf{0.24} & \textbf{0.06} & \textbf{542} & \textbf{36} \\
    \bottomrule
    \end{tabular}
\end{table}

\subsubsection{\textbf{Runtime analysis}} 
Finally, we report a runtime analysis in Table~\ref{tab:runtime} to confirm that the system operates in real time. Across all experiments, \textit{ivS-Graphs} consistently maintained processing rates between 21.8 and 24.9\,FPS (averaging 23.3\,FPS), staying above the 20\,FPS real-time threshold. While this is lower than the 29.3\,FPS average achieved by \textit{ORB-SLAM3}~\cite{orb3}, the difference is primarily due to the additional computational cost of the \textit{Scene Segmentor} and the subsequent structural element extraction. 
\add{The \textit{Match} and \textit{Opt.} columns of Table~\ref{tab:runtime} isolate the BIM-specific cost.
\addtwo{Match performs an unindexed linear scan over the BIM wall database, giving $O(N)$ complexity, but its cost (0.07\,ms, our highest measured cost) is negligible when compared with Opt.'s cost ($\sim$540\,ms average), several orders of magnitude larger.} \changetwo{Opt.'s per-call cost is dominated by newly added associations and the number of poses being corrected; established associations contribute little, as their residuals have already converged.}{Opt.'s per-call cost is driven by trajectory length rather than wall count: office1-1, the longest of these four sequences, has the highest cost despite fewer walls than CS1 or CS2-Mid.} 
\addtwo{Neither cost affects the system's real-time frame rate reported above, since Match and Opt. run in the local mapping thread, decoupled from tracking. For substantially larger buildings, more complex optimization strategies may become necessary. Hierarchical strategies, such as per-room or per-floor optimization~\cite{bavle2025s}, could be implemented, but this is out of the scope of this work.}}
Additionally, the initial alignment is performed in only 0.24\,ms,
which is negligible for system startup.

\subsubsection{\textbf{Limitations}}
\addtwo{Wall detection is bounded by the depth-sensing range of the RGB-D camera feeding the \textit{Scene Segmentor}, which extracts the detected walls $\mathbf{W_S}$ only within this range, with accuracy degrading near the limit. The sensor should therefore be selected according to the depth range required by the target environment.}
\addtwo{These same detection limits bear directly on initialization: if the two intended walls go undetected (occluded, out of range, or otherwise not mapped) or the operator specifies the wrong wall IDs, the acceptance checks ($\tau_\perp$, $\tau_{init}$, Table~\ref{tab:parameters}) are not satisfied, initialization does not occur, and the system runs as the SLAM backbone. If a wrong pair coincidentally satisfies these checks, the resulting alignment is wrong, producing unexpected results.}
\section{Conclusions and Future Work}
\label{sec:conclusions}
We presented \textit{ivS-Graphs}, the first visual SLAM system to integrate BIM priors for construction environments, producing drift-bounded \textit{as-built} maps with correspondences to the \textit{as-planned} design. Experimental results show a 25.23\% ATE reduction and 7.14\% map RMSE improvement over the visual SLAM backbone, with the largest gain on longer trajectories where drift accumulates more.
The system stays robust under 30\% BIM wall omission (5.3\% error increase) and handles geometric deviations via the uncertainty-aware back-end.

In future work, we aim to integrate additional BIM elements, such as windows, columns, and pipes, to further enhance the alignment between the \textit{as-built} and \textit{as-planned} maps. \changetwo{Building on this, we plan to investigate}{We also plan to investigate} \changetwo{alternative initialization strategies that leverage these additional semantic cues to provide more informative initial alignment, reducing the need for the manual initial wall correspondence}{automatically detecting an incorrect initialization}.


\bibliographystyle{IEEEtran}
\bibliography{root}

\end{document}